# High Efficient Reconstruction of Single-shot T2 Mapping from OverLapping-Echo Detachment Planar Imaging Based on Deep Residual Network

Congbo Cai[1], Yiqing Zeng[1], Chao Wang[1], Shuhui Cai[2], Jun Zhang[2], Zhong Chen[2], Xinghao Ding[1,*] and Jianhui Zhong[3]

[1]*Department of Communication Engineering, Xiamen University, Xiamen 361005, China*

[2]*Department of Electronic Science, Fujian Provincial Key Laboratory of Plasma and Magnetic Resonance, Xiamen University, Xiamen 361005, China*

[3] *Department of Imaging Sciences, University of Rochester, Rochester, NY 14642, USA, the Center for Brain Imaging Science and Technology and Collaborative Innovation Center for Diagnosis and Treatment of Infectious Diseases, Zhejiang University, Hangzhou 310007 , China*

Running title: High efficient reconstruction of OLED imaging based on deep residual network

Word count: 4462

Figures: 7

Grant sponsor: National Natural Science Foundation of China; Grant numbers: 81171331, 81671674, 11474236, 61571382 and 81301277.

---

[1] Correspondence to: Xinghao Ding, Ph.D., Department of Communication Engineering, Xiamen University, Xiamen, 361005, China. E-mail: dxh@xmu.edu.cn.

## ABSTRACT

**Purpose:** An end-to-end deep convolutional neural network (CNN) based on deep residual network (ResNet) was proposed to efficiently reconstruct reliable $T_2$ mapping from single-shot OverLapping-Echo Detachment (OLED) planar imaging.

**Methods:** The training dataset was obtained from simulations carried out on SPROM software developed by our group. The relationship between the original OLED image containing two echo signals and the corresponded $T_2$ mapping was learned by ResNet training. After the ResNet was trained, it was applied to reconstruct the $T_2$ mapping from simulation and in vivo human brain data.

**Results:** Though the ResNet was trained entirely on simulated data, the trained network was generalized well to real human brain data. The results from simulation and in vivo human brain experiments show that the proposed method significantly outperformed the echo-detachment-based method. Reliable $T_2$ mapping was achieved within tens of milliseconds after the network had been trained while the echo-detachment-based OLED reconstruction method took minutes.

**Conclusion:** The proposed method will greatly facilitate real-time dynamic and quantitative MR imaging via OLED sequence, and ResNet has the potential to reconstruct images from complex MRI sequence efficiently.

.

# INTRODUCTION

Though traditional magnetic resonance imaging (MRI) can provide excellent tissue contrast, it is usually qualitative or “weighted” measurement (1). MR signal is not only influenced by several intrinsic contrast mechanisms, but also affected by different imaging system (2). Quantitative MR mapping can provide quantitative information for characterizing specific tissue properties (3), and help to remove unrelated influences on disease diagnosis and make the comparison available across different sites, MRI protocols and scanner vendors. Quantitative $T_2$ mapping has drawn more and more attention in clinical MRI (4,5). Neurologic and psychiatric diseases demonstrate variations in $T_2$ values in brain, such as stroke (6), multiple sclerosis (7) and epilepsy (8). Myocardial $T_2$ mapping is thought to be meaningful in diagnosis of acute myocardial conditions associated with myocardial edema (9).

However, long data acquisition time usually hinders the practical applications of $T_2$ mapping, especially in the cases that requires high temporal resolution or with irregular motions. In our early work, a novel single-shot $T_2$ mapping sequence based on SE-EPI acquisition scheme (10) was proposed. Two overlapped echo signals with different $T_2$ weighting were obtained simultaneously by using two excitation pulses with small flip-angle and corresponding echo-shifting gradients to shift the echo centers respectively. Detachment algorithm based on structure similarity constraint was proposed to separate the two echo signals. The method called single-shot $T_2$ mapping through overlapping-echo detachment (OLED) planar imaging shows good performance in obtaining reliable $T_2$ mapping efficiently. The total scanning time for OLED sequence only increases by about 10% compared to conventional SE-EPI sequence. However, due to the highly non-linear mapping process of OLED imaging, the efficiency of reconstruction based on priori constraints is still limited and the reconstruction is rather slow (in minutes), which hinders its application in real-time imaging.

Deep learning, a family of algorithms for efficient learning of complicated dependencies between input data and outputs by propagating a training dataset through several layers of hidden units, has shown explosive popularity in recent years with the availability of powerful GPUs (11). Deep neural network architectures, convolutional neural network (CNN) in particular, are becoming more and more popular in medical imaging analysis for various problems including classification (12), detection (13) and segmentation (14). It has already been demonstrated that CNN outperforms sparsity-based methods in super-resolution reconstruction (15), not only for its quality but also in terms of the reconstruction speed. In this work, we will explore the application of CNN in MR parameter mapping reconstruction and investigate whether it can exploit data redundancy through learned representations. Despite the popularity of CNN, there are still only a little preliminary research on CNN-based MR image reconstruction, hence the applicability of CNN to this problem is yet to be qualitatively and quantitatively assessed in detail.

As an outstanding representative of deep neural network, deep Residual Network (ResNet) can mitigate the vanishing/exploding gradient problem during training for a deep network (16). Therefore, it can quickly accelerate the training of ultra-deep neural network, and the results of training can be more accurate. In the present work, an end-to-end deep convolutional network based on deep residual network was demonstrated to be able to provide better reconstruction of $T_2$ mapping with much faster reconstruction speed and better quality compared to the echo-detachment-based method.

## MATERIALS AND METHOD

### *A. Problem Formulation*

The OLED sequence is shown in Fig. 1. Two excitation pulses with a same small flip angle $\alpha$ are used to sequentially produce two echo signals with different evolution time. However, in practice, three echo signals with different $T_2$ weighting are obtained in the same k-space within a single shot:

$$\begin{cases} S_1(TE_1) = \frac{1}{2}\int_{\vec{r}} \rho(\vec{r})\left|\sin\alpha\cos\alpha\right|\left(1-\cos\beta\right)e^{-TE_1/T_2(\vec{r})}d\vec{r}, & \text{first spin echo} \\ S_2(TE_2) = \frac{1}{4}\int_{\vec{r}} \rho(\vec{r})\left|\sin\alpha\right|\left(1+\cos\alpha\right)\left(1-\cos\beta\right)e^{-TE_2/T_2(\vec{r})}d\vec{r}, & \text{second spin echo} \\ S_3(TE_1) = \frac{1}{4}\int_{\vec{r}} \rho(\vec{r})\left|\sin\alpha\right|\left(1-\cos\alpha\right)\left(1-\cos\beta\right)e^{-TE_1/T_2(\vec{r})}d\vec{r}, & \text{double-spin-echo} \end{cases} \quad [1]$$

where $\rho(\vec{r})$ denotes spin density at position $\vec{r}$, $\alpha$ is the flip angle of excitation pulse, and $\beta$ is the refocusing pulse. The first spin echo is produced by the second excitation pulse and the second spin echo is produced by the first excitation pulse, while the double-spin-echo is refocused by the last two RF pulses. According to Eq. [1], the relationship between the overlapped echo signals and the $T_2$ value is nonlinear. An echo-detachment-based method based on the sparsity constraint and structure similarity constraint has been utilized to separate the overlapped echoes and calculate the corresponding $T_2$ mapping according to the $T_2$ relaxation attenuation formula. To separate the overlapped echoes, the following energy function is minimized:

$$\{\mathbf{x}_1, \mathbf{x}_2\} = \underset{\mathbf{x}_1,\mathbf{x}_2}{\arg\min} \begin{bmatrix} \left\|\mathbf{x}_0 - \mathbf{x}_1 \cdot e^{i\varphi_1(\mathbf{r})} - \mathbf{x}_2 \cdot e^{i\varphi_2(\mathbf{r})}\right\|_2^2 + \lambda_1 \left\|\mathbf{M}\cdot\nabla\mathbf{x}_1\right\|_1 \\ +\lambda_2 \left\|\mathbf{M}\cdot\nabla\mathbf{x}_2\right\|_1 + \lambda_3 \left\|\mathbf{M}\cdot\nabla\left(\mathbf{x}_1 - \kappa\mathbf{x}_2\right)\right\|_1 \end{bmatrix}, \quad [2]$$

where $\mathbf{x}_0$ is the original image including the first and second echo signals, $\mathbf{x}_1$ and $\mathbf{x}_2$ are separated images from the first and second echo signals, respectively, $\varphi_1(\mathbf{r})$ and $\varphi_2(\mathbf{r})$ are linear phase ramps of the first and second images respectively. $\lambda_1$, $\lambda_2$, and $\lambda_3$ are Lagrange multipliers for adjusting constraint weights. $\|\cdot\|_1$ stands for the $\ell_1$ norm, which promotes the sparsity of each image, $\|\cdot\|_2$ stands for the $\ell_2$ norm. $\nabla$ denotes the gradient operator. $\kappa$ is a scaling factor, and **M** is an edge-weighting matrix. Eq. [2] can be solved by using iterative reconstruction method. The detail of the echo-detachment-based method can be found in (10).

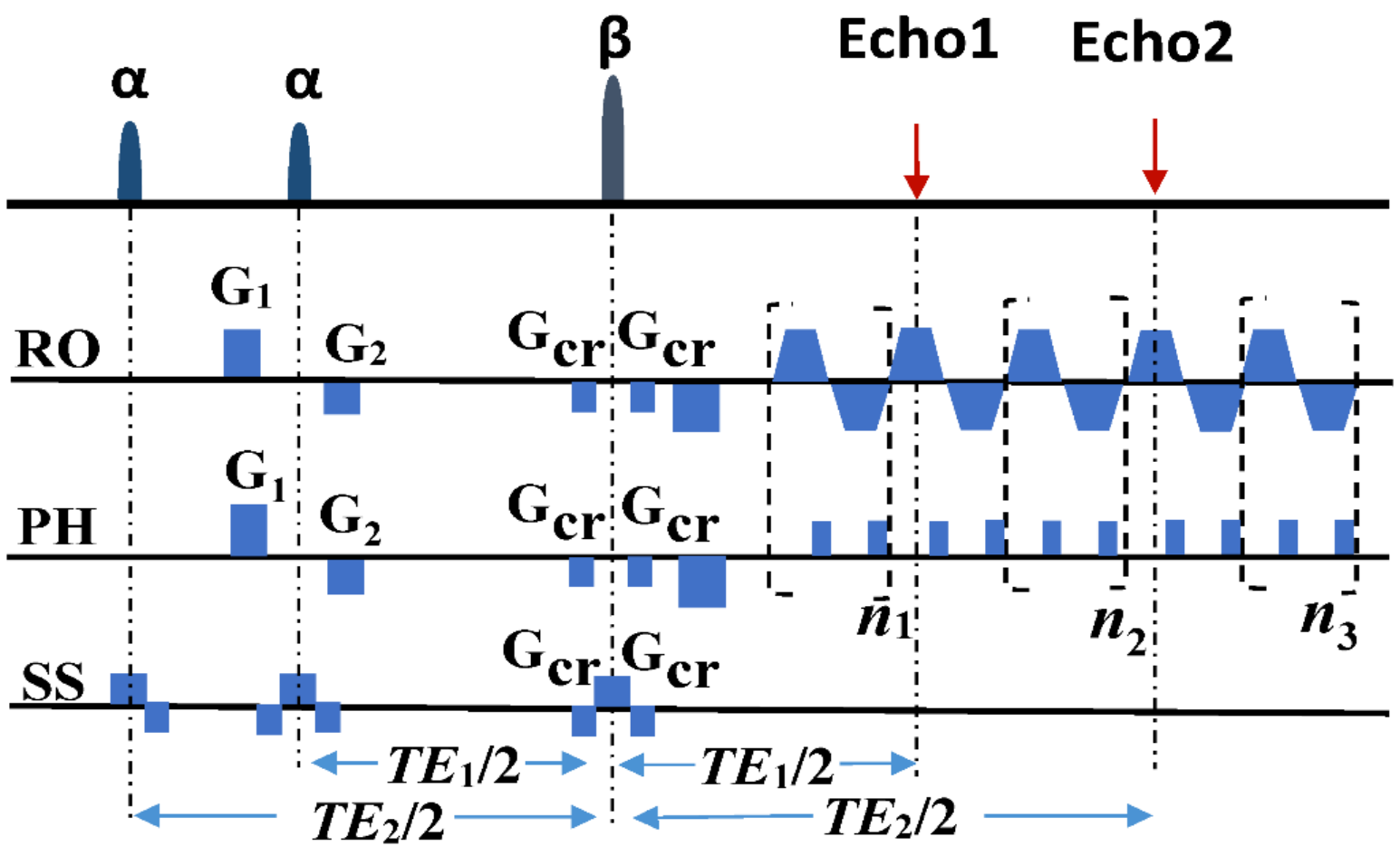

**FIG. 1.** Single-shot OLED sequence. The first two RF pulses are excitation pulses with flip angle α, the third RF pulse is refocusing pulse with flip angle β. $G_1$ and $G_2$ are first and second echo-shifting gradients, and $G_{cr}$ represents crusher gradients along three directions. $n_1+n_2+n_3+4=N$, where $N$ is the echo number of OLED.

In this work, deep learning method adjusts the parameters of a multilayer neural network such that the outputs of the network will approximate the target $T_2$ mapping when the original OLED image is input. After the ResNet is trained, it is applied to reconstruct the real experimental data.

### *B. Network Architectures*

The proposed end-to-end framework for reconstructing the $T_2$ mapping from OLED sequence is illustrated in Fig. 2. For traditional CNN, with the increasing of neural network depth, the accuracy will rise first and then reach saturation, and finally the accuracy may decrease. It is not about over-fitting, because the training error also increases. Assuming that H(x) is an underlying mapping to be fit by some stacked layers, and x denotes the inputs to the first layer, if the input is passed directly to the output, the stacked nonlinear layers will fit another mapping of F(x) = H(x) – x, that is called residual function. Thus the original function becomes F(x) + x, which can be realized by feed forward neural networks with "shortcut connections"( 17 ). The shortcut connections are shown in Fig. 2. The traditional convolutional layers in the process of information transmission will loss information more or less. ResNet solves this problem by directly transferring the input information to the output to protect the integrity of the information. Since the entire network only needs to learn the difference between input and output, the learning objectives and difficulty can be simplified. Here we modified the popular ResNet model to be suitable for image regression problems (18) (as shown in Fig. 2).

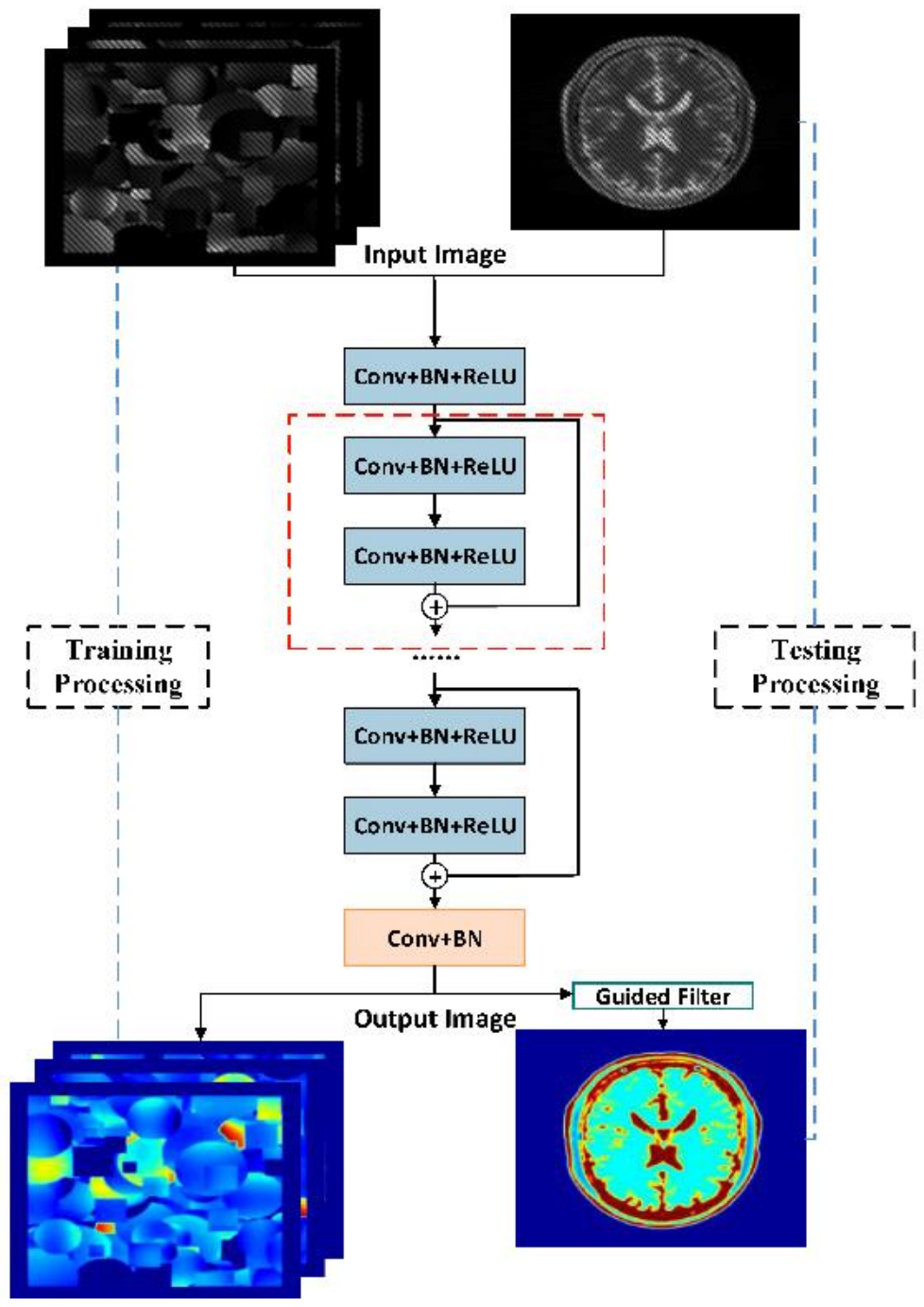

**FIG. 2.** The adopted framework based on ResNet for reconstructing $T_2$ mapping from OLED sequence. The red dashed box illustrates the residual unit.

The input of our OLED-reconstruction system is an overlapping-echo signal image X and the output is an approximation to the $T_2$ image Y. So we define the following objective function:

$$\mathcal{L} = \frac{1}{N}\sum_{i=1}^{N} \| f\left(\mathbf{X}^i, \mathbf{W}, \mathbf{b}\right) - Y^i \|_F^2 \quad [3]$$

where $N$ is the batch number of training images, $f(\cdot)$ is the deep residual network, **W** and **b** are network parameters that need to be learned. The basic network structure can be expressed as:

$$\begin{aligned}
&\mathbf{X}^0 = \mathbf{X} \\
&\mathbf{X}^1 = \sigma(\mathrm{BN}(\mathbf{W}^1 * \mathbf{X}^0 + \mathbf{b}^1)) \\
&\mathbf{X}^{2l} = \sigma(\mathrm{BN}(\mathbf{W}^{2l} * \mathbf{X}^{2l-1} + \mathbf{b}^{2l})) \\
&\mathbf{X}^{2l+1} = \sigma(\mathrm{BN}(\mathbf{W}^{2l+1} * \mathbf{X}^{2l} + \mathbf{b}^{2l+1})) + \mathbf{X}^{2l-1} \\
&\mathbf{Y}_{\text{approx}} = \mathrm{BN}(\mathbf{W}^L * \mathbf{X}^{L-1} + \mathbf{b}^L)
\end{aligned} \quad [4]$$

where $l = 1, \ldots, \frac{L-2}{2}$, and $L$ is the total number of layers, $*$ indicates the convolution operation, $\mathbf{W}$ represents weights and $\mathbf{b}$ represents biases, $\mathrm{BN}(\cdot)$ indicates batch normalization (19) to alleviate internal covariate shift, $\sigma(\cdot)$ is a Rectified Linear Unit (ReLU) for non-linearity (20). In the network, all pooling operations are removed to preserve the spatial information.

For the first layer, filters with size $c \times s_1 \times s_1 \times a_1$ are used to generate $a_1$ feature maps. Here, *s* represents filter size and *c* denotes the number of image channels. In the present work, our network takes in a two-channel image where each channel stores real/imaginary parts of the complex input image. For layers 2 to $L$-1, the filters have a size of $a_1 \times s_2 \times s_2 \times a_2$. For the last layer *L*, filters with size $a_2 \times s_3 \times s_3 \times 1$ are used to estimate the output of the network. It is worth noting that the output image only contains $T_2$ values, so for the last layer *L* the number of image channel is 1. In the reconstruction, a guided filter (21) is applied to remove the unexpected noises in the real experimental data, where the guided image is the output image itself with the radius of the guided filter 15 (18).

### *C. Training Dataset*

In this work, the training dataset including OLED images and corresponding $T_2$ mapping was obtained from simulation via SPROM software developed by our group (22). To cope with various complex textures possibly occurring in the real data reconstruction, random $T_2$ and proton density templates were produced. To enhance the robustness of CNN against non-ideal experimental conditions (e.g. the echo-shifting gradients may not be ideal in experiments), some sequence parameters were varied slightly and randomly. To enhance the robustness of reconstruction against noises, Gaussian white noise with a signal-to-noise (SNR) of 115.3 was added to the simulated OLED images according to $\mathrm{SNR} = 20\lg\frac{\mu_s}{\delta_n}$, where $\mu_s$ was the mean value of the signal intensities and $\delta_n$ was the standard deviation of added noises. A dataset containing 100 pairs of original OLED images and corresponding $T_2$ images was produced for ResNet training.

*D. Data Acquisition*

Numerical simulations and in vivo human brain experiments were performed to validate ResNet reconstruction method for OLED. Experiments were performed on a whole-body MRI system at 3T (MAGNETOM Trio TIM, Siemens Healthcare, Erlangen, Germany). The study protocol was approved by the local research ethics committee and informed consent was obtained before examination. Simulations were carried out with the SPROM software developed in-house on a personal computer (Intel Core i7, 32GB memories, 64 bit Windows 10 operation system), and the data pre-processing was performed using MATLAB R2010b software (Mathworks, Natick, MA, USA). It should be pointed out that the same simulation data and in vivo human brain data were used in our previous work (10).

Numerical simulation

A numerical BrainWeb phantom (including $T_2$, $T_1$ and proton-density template) was constructed (23). A matrix size of 512×512 grids and a field of view (FOV) of 22×22 $cm^2$ were used for the two-dimensional model in the simulation. The OLED sequence with the flip angle $\alpha$ = 45° was used. The durations of echo train for the OLED and EPI sequences were about 90 ms. For the OLED sequence, $\Delta TE \approx 46$ ms and a Gaussian white noise was added with the SNR = 70.1 dB.

Human Brain Experiments

The OLED and SE images were acquired from two healthy volunteers (age 26~41 years). The experimental parameters were the same as those used in the simulation. Four different TEs (35 ms, 50ms, 70 ms, and 90 ms) were utilized for conventional SE sequence with total scan time of 17 min. For OLED, the flip angle of excitation pulses was 50°, and $\Delta TE \approx 45$ ms was used. The slice thickness was 4 mm for all sequences.

*E. Training*

The ResNet had 8 parameter layers and stochastic gradient descent (SGD) was used to minimize the objective function in Eq. [3] with the weight decay of $10^{-8}$, momentum of 0.9 and mini-batch size of 16. We started with a learning rate of 0.1, divided the rate by 10 at 32k and 64k iterations, and terminated training at 100k iterations. The filter size was 3×3 and filter number was 64. No augmentation was used. The batch normalization (BN) was adopted right after each convolution and before activation.

As demonstrated in our early work, the original k-space data contain three echoes, while the double-spin-echo should be removed by a simple Gauss filter before the training and testing. We randomly selected 90 images to train our network from which 64×64 patch pairs were randomly cropped and the remaining 10 images were used to test the network. Caffe software package was used to train our network (24). It took approximately 2.5 hours to train our network.

## RESULTS

To evaluate the trained network, the $T_2$ mappings reconstructed from ResNet were compared with those from the echo-detachment-based method for numerical simulation and *in vivo* human brain.

### *A. Numerical Simulation*

The results of numerical simulation are shown in Fig. 3. From Fig. 3(a), we can see that the original OLED amplitude image has obvious streaks, since the image is composed of two images with different linear phase ramp. Comparing Figs. 3(c), 3(d) and 3(e), one can see that the reconstructed $T_2$ mapping from both echo-detachment-based method and deep learning method agree well with the reference $T_2$ mapping. Fig. 3(b) shows the image obtained with conventional single-shot EPI sequence using the same parameters as those for OLED sequence. The $T_2$ profiles along the red trace from deep learning method, echo-detachment-based method and reference $T_2$ mapping were drawn in Fig. 3(f). From Fig. 3(f), we can see that deep learning method can obtain higher accuracy compared with echo-detachment-based method, especially in the tiny CSF (cerebrospinal fluid) regions, which usually have longer $T_2$ value. A region in $T_2$ mapping is expanded for better comparison. We can see that deep learning method can provide better reconstruction resolution and SNR than the echo-detachment-based method. For the echo-detachment-based method, there are some significant mismatches at boundaries of different tissues with abrupt variation of $T_2$ values. The reconstructed $T_2$ mapping from deep learning method agrees better with the reference $T_2$ mapping not only at the boundaries of different tissues but also in the smooth regions. It should be noted that both the relatively low image resolution (128×128 for OLED versus 256×256 for reference $T_2$ mapping) and point spread function of $T_2$ decay with long echo train for OLED still cause some blurring effects in the ResNet reconstruction result compared with the reference $T_2$ mapping.

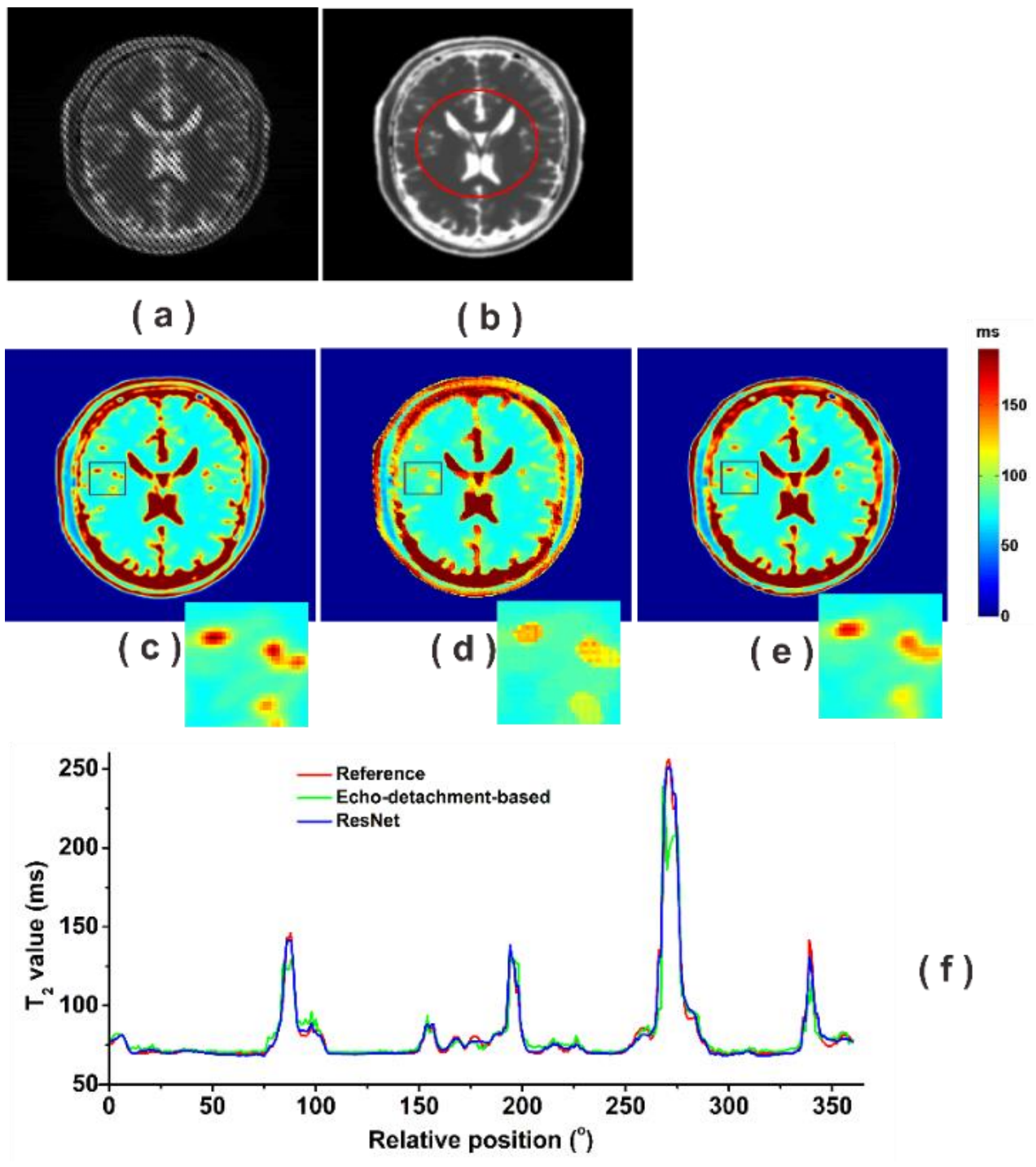

**FIG. 3.** Simulated MR images (22 × 22 $cm^2$) obtained from single-shot OLED sequence. The total scan time is 147.6 ms (for OLED) and 135.9 ms (for EPI) with acquisition matrix = 128 × 128, *sw* = 1.42 kHz/pixel. The vertical axis is phase-encoded and the horizontal axis is frequency-encoded if not specified. Each image matrix is expanded to 256 × 256 through zero-padding before fast Fourier transform. (a) Original OLED amplitude image containing two echo signals; (b) Spin-echo EPI image; (c) Reference $T_2$ mapping; (d) $T_2$ mapping reconstructed from (a) using echo-detachment-based method; (e) $T_2$ mapping reconstructed from (a) using ResNet; (f) $T_2$ value traces of (c), (d) and (e) along the red circle line in (b). The trace starts at the three-o'clock position of the circle and runs anticlockwise.

## B. *In Vivo Human Brain*

The results of three different slices from human brain are shown in Fig. 4.

Compared to the echo-detachment-based method, we can see that the deep learning method can provide a result with lower noises and higher resolution. Some small artifacts are still obvious from the echo-detachment-based method even though the total variation (TV) regularization has been applied, while ResNet method can remove the artifacts and preserve details better.

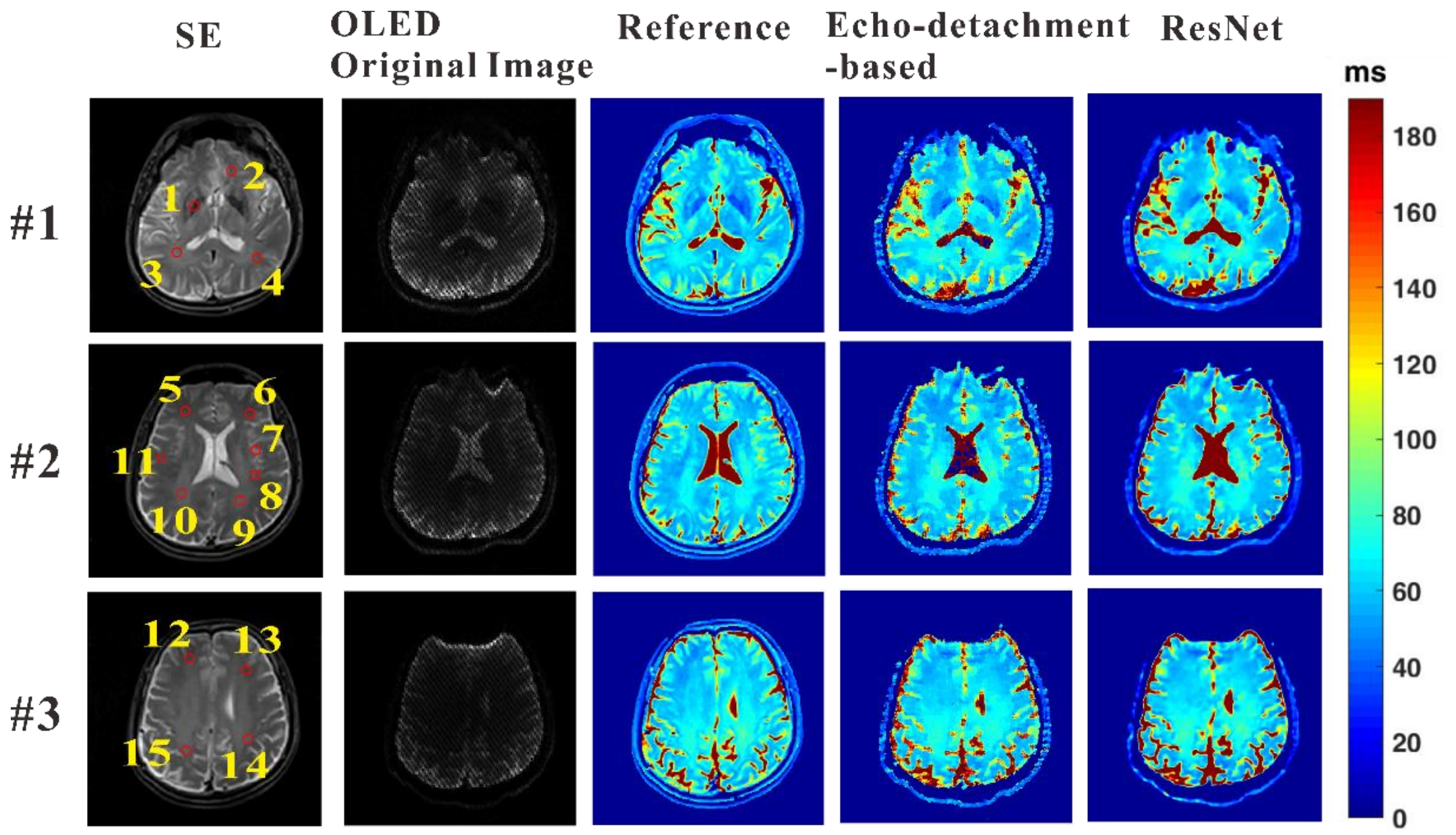

**FIG. 4.** MR images (22×22 $cm^2$) of human brain. Original OLED images contain two echo signals. Fifteen ROIs were marked with red circles and numbered in conventional SE images. The gray scales in 1st, and 2nd columns were normalized respectively.

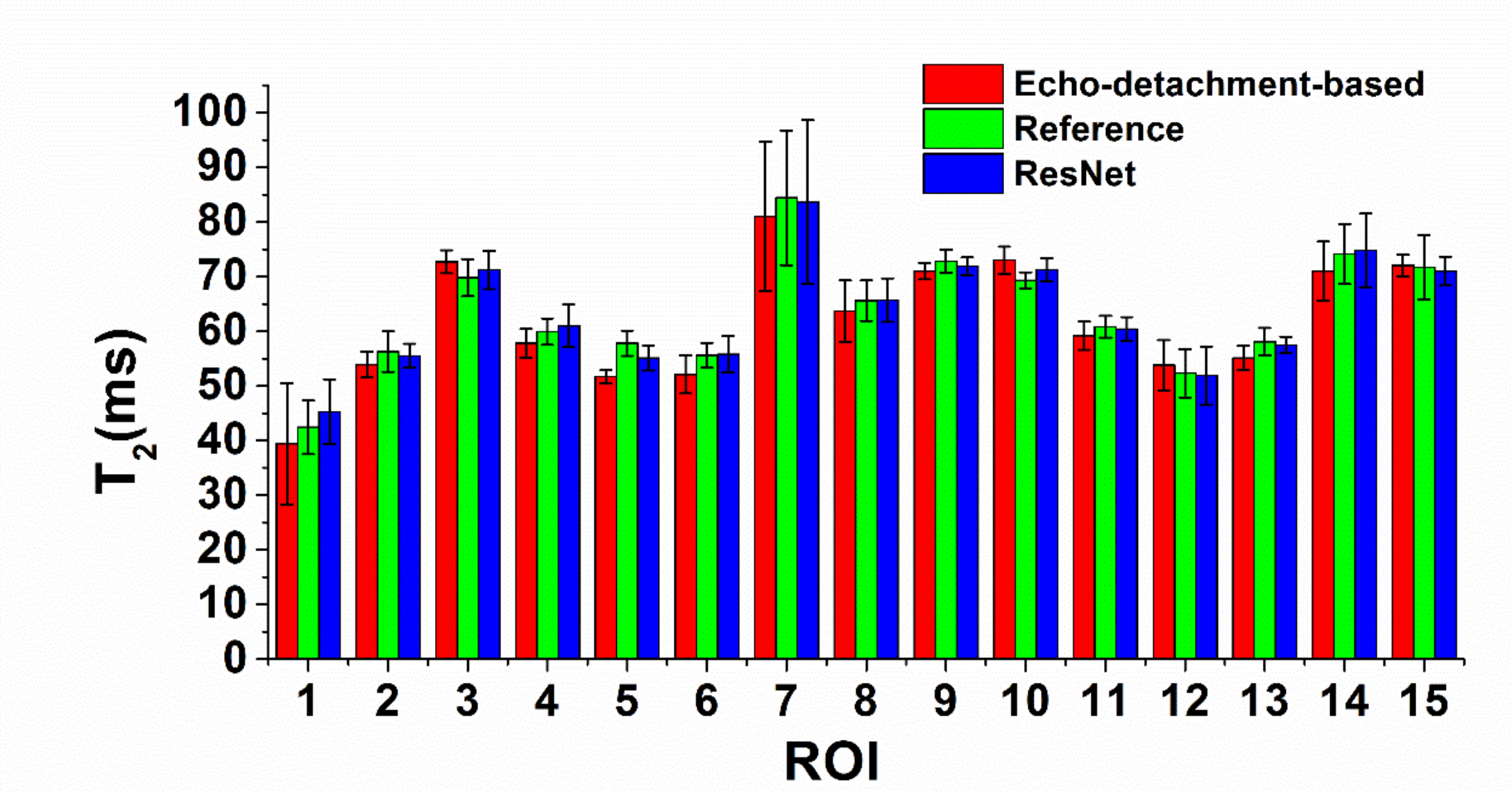

**FIG. 5.** Mean $T_2$ values and standard deviations (shown as vertical bars) for 15 ROIs marked in Fig. 4 for SE (reference), echo-detachment-based and ResNet methods.

Fig. 5 illustrates the mean $T_2$ values and standard deviations for echo-detachment-based method and deep learning method. The values were calculated from 15 regions-of-interest (ROIs) marked in Fig. 4. The 15 ROIs were chosen deliberately to represent different regions of the brain, including white matter, gray matter and deep gray matter with quite different $T_2$ values. The $T_2$ mappings reconstructed by echo-detachment-based method and deep learning method from OLED sequence were all co-registered to reference $T_2$ mappings obtained from multi-scan SE sequence using the FNIRT tool (FMRIB's Nonlinear Image Registration Tool) of FSL (Oxford, UK). We can see that both echo-detachment-based and ResNet results agree quite well with SE in general. However, the ResNet result is better in most ROIs. ROI5 shows the largest deviation between the echo-detachment-based and SE $T_2$ mappings ($\Delta T_2$ = 6.2 ms, 10.7%), while the largest deviation between ResNet and SE $T_2$ mappings appears in ROI1 ($\Delta T_2$ = 3.0 ms, 7.0%).

## DISCUSSION

To reconstruct reliable $T_2$ mapping from single-shot OLED sequence, an echo-detachment-based method based on structure similarity constraint has been proposed to separate the two overlapped signals first and then calculate corresponding $T_2$ values according to the signal evolution formula under $T_2$ relaxation attenuation. However, the echo-detachment-based method separates two overlapped echoes using low-level image features. When the structure and orientation of an object are similar to those of modulation streaks, it would be hard for the echo-detachment-based method to simultaneously separate the overlapped signal and preserve the object's structure. From the expanded region in Fig. 3d, we can see that some residual artifacts (lattice-shaped) cannot be removed even though the TV regularization is used. These artifacts are corresponding to the modulation streaks.

The detachment of two overlapped echo signals is not necessary, because only the reconstructed $T_2$ mapping is desired. In this work, an end-to-end reconstruction method based on ResNet was proposed to directly reconstruct the $T_2$ mapping from single-shot OLED sequence. ResNet is effective at capturing high-level image features about phase modulation and object structure. The results from numerical simulation and in vivo human brain measurements show that the ResNet reconstruction method performs well in reconstructing $T_2$ mapping within tens of milliseconds, far shorter than minutes in echo-detachment-based method. It should be noted that although we train the network with simulation data, the learned network generalizes well to real data.

A qualified training dataset is vital for CNN-based reconstruction method. To the best of our knowledge, all the CNN-based methods in MRI are trained with real experimental dataset. It is not surprising because real experimental condition is complex, and simulation may not take all imperfect factors into account, which may deviate the training data from real experimental data and make the reconstruction of real data failed. However, there are also limitations in using real experimental data as training dataset. Firstly, we must acquire huge experimental data, which may take much experimental time and scanning cost. Secondly, the experimental conditions may vary under different imaging objects, scanners or even at different scan time, whereas the experimental training dataset may not cover all these conditions. Finally, the training label may be not available for some sequences and a different imaging sequence has to be selected as reference sequence. For example, the target data from ultrafast imaging sequence need to be reconstructed, while the reference data can only be achieved from a relatively slow imaging sequence. Different sequences usually bring in different artifacts, such as distortions from inhomogeneous field, chemical shift artifacts from fat signal, and eddy effects, and they may also have different contrast mechanism. Therefore, when we use the images from multi-scan imaging sequence as the training label for single-shot imaging sequence and train the network, the differences between these two sequences may cause serious problem if single-shot data is reconstructed by this network.

In this paper, an 8-layer ResNet was used. A deeper network may have greater fitting capacity due to additional nonlinearities and better quality of local optima (25, 26). However, it also needs more computation and may increase the risk of overfitting. The reconstructed $T_2$ values inside the marked ROIs under different ResNet layers are shown in Fig. 6. We can see that adding additional layers to the architecture does not provide any meaningful performance improvement. The results under 8-layer, 11-layer and 17-layer ResNet are similar though the latter needs much more training time. In some ROIs (e.g. ROI5), 11-layer and 17-layer ResNet even give slightly worse results with bigger deviation from reference $T_2$ values. Therefore, 8-layer ResNet should be enough for efficient reconstruction of $T_2$ mapping from the OLED sequence.

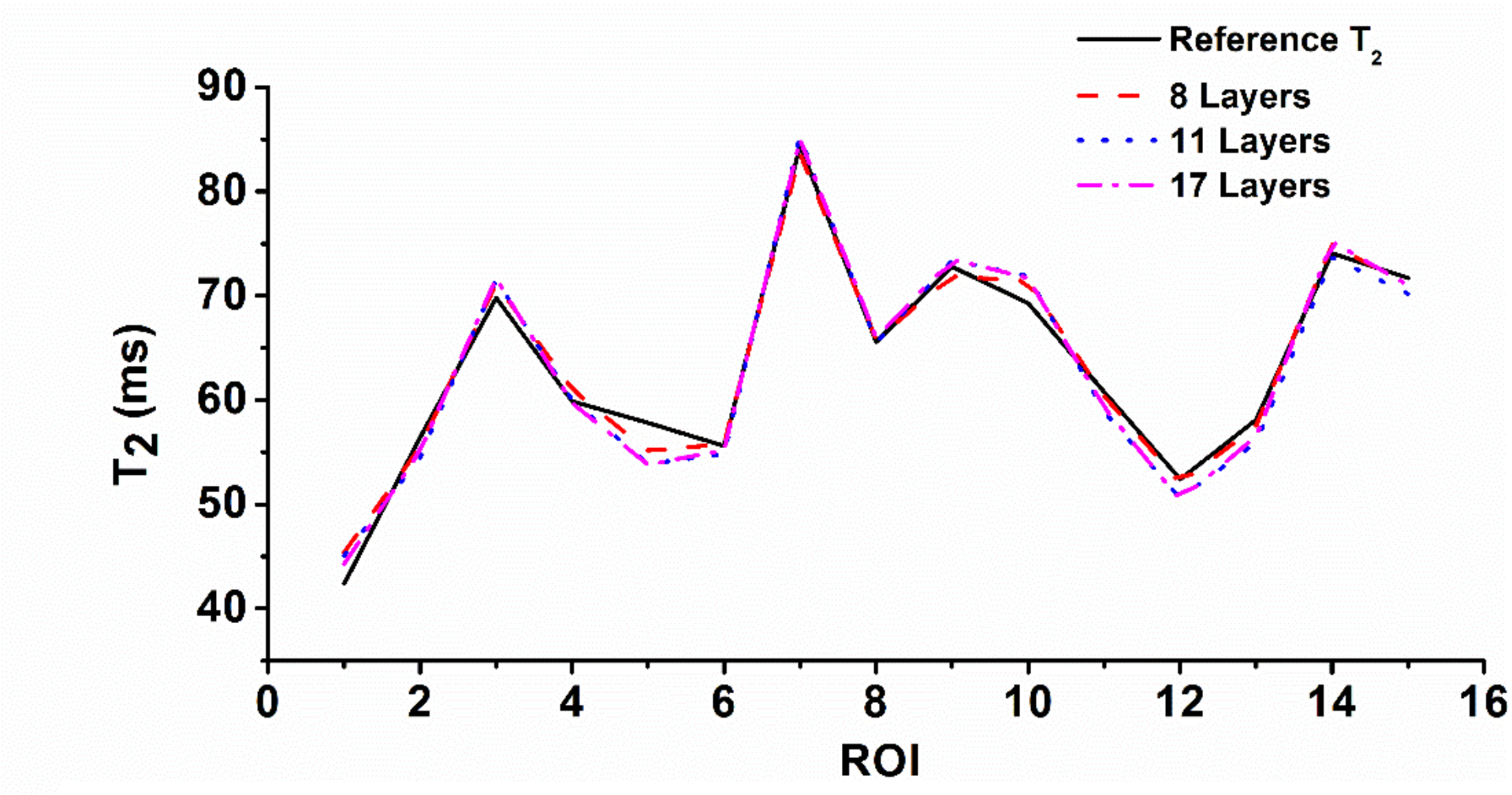

**FIG. 6.** Reconstructed $T_2$ values inside the marked ROIs under different ResNet layers.

In real experiments, ideal experimental conditions are hardly satisfied, e.g. the gradients may not be ideal. Non-ideal experimental conditions may crush the CNN-based reconstruction if they are not considered in the training. For the single-shot OLED sequence, non-ideal echo-shifting gradients will affect the signal modulation of OLED images, and thus degrade the reconstruction quality of CNN if this non-ideal factor is not considered in the training. In Fig. 7, a numerical human brain was used to verify the sensitive of ResNet reconstruction method to non-ideal echo-shifting gradients. Multiple-sequence training dataset represents the data produced by the same OLED sequence but with slightly different echo-shifting gradients along frequency and phase dimensions, while single-sequence training dataset represents the data produced by one sequence with ideal echo-shifting gradients. From Fig. 7, we can see that the multiple-sequence training dataset can provide more robust results compared to the single-sequence training dataset under different non-ideal echo-shifting gradients. The single-sequence training dataset gives biggest $T_2$ value deviation (6.8 ms, about 9.7%) when the test data is produced by 10% deviation of echo-shifting gradients along both dimensions, while the biggest deviation of $T_2$ value from multiple-sequence training dataset is only 1.1 ms (about 1.6%).

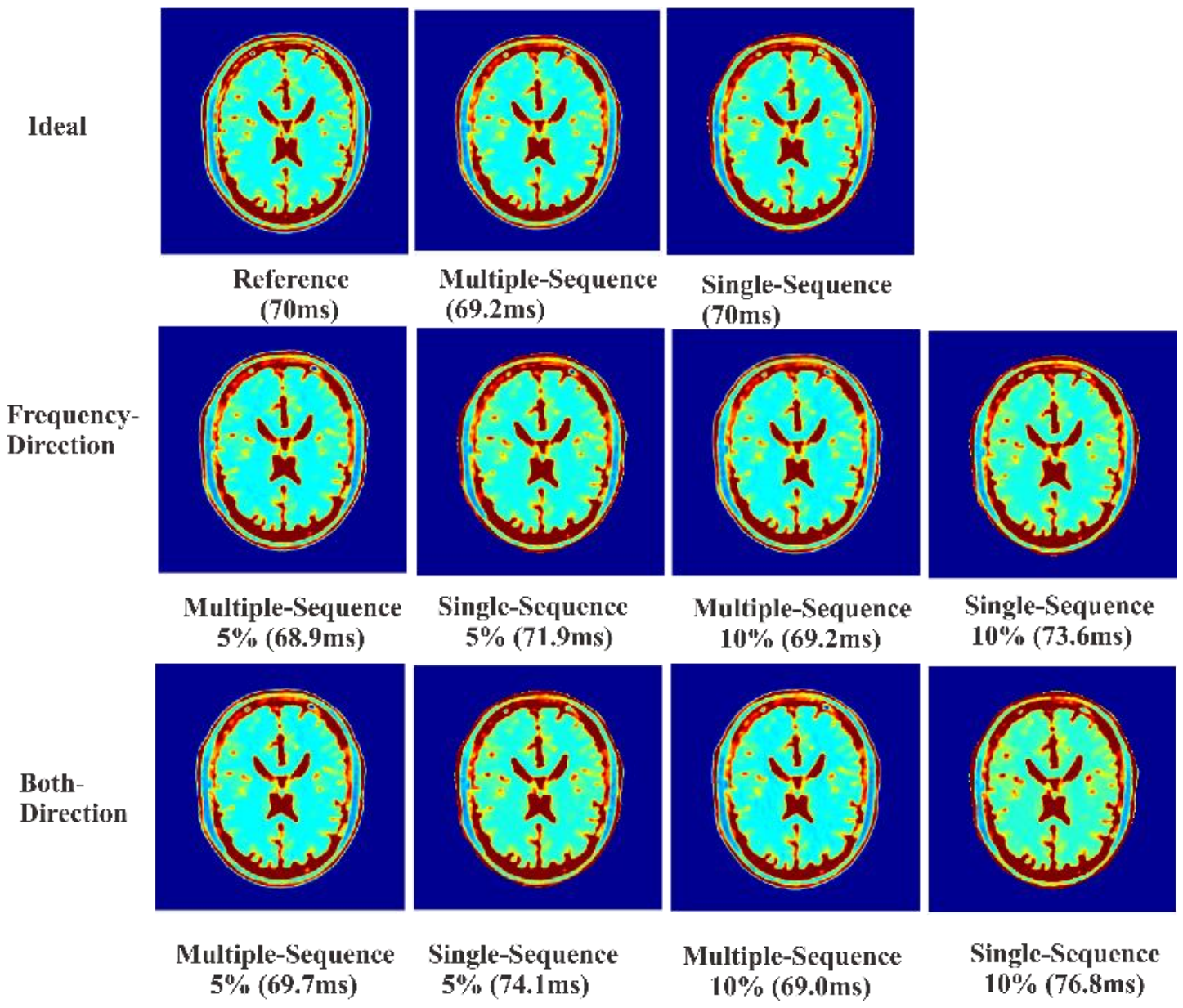

**FIG. 7.** Reconstructed $T_2$ mappings of numerical human brain under different non-ideal echo-shifting gradients and different training datasets. Frequency-direction / Both-direction means the echo-shifting gradients of test data are varied (with 5% and 10% of gradient amplitude) along frequency dimension / both frequency and phase dimensions. The mean $T_2$ values are obtained from a relatively smooth small region.

## CONCLUSION

This paper shows that deep learning method can help to improve the reconstruction of single-shot $T_2$ mapping from OLED sequence in image quality and reconstruction speed. Simulated dataset was used to train the network, and the trained network was generalized quite well to real experimental results. The pure simulated data are used to train the deep network and reconstruct the MR images. This implies that the deep learning method will open a new door to the reconstruction of images from complex MRI sequence where traditional optimization-based reconstruction method may be helpless, and the simulated training dataset will help to solve the difficulties in obtaining enough qualified training dataset from experiments. Our work may help to promote the applications of deep network in MR image reconstructions.